\pdfoutput=1

\documentclass[11pt]{article}

\usepackage{acl}

\usepackage{stfloats}
\usepackage{ntheorem}   
\usepackage{lipsum}     

\theoremstyle{plain}
\theoremheaderfont{\bfseries}

\usepackage{amssymb}
\usepackage{pifont}
\usepackage{times}
\usepackage{latexsym}
\usepackage{helvet}
\usepackage{courier}
\usepackage{amsmath}
\usepackage{amsfonts}
\usepackage{amssymb}
\usepackage{multirow}
\usepackage{multicol}
\usepackage{graphicx}
\usepackage{subfigure}
\usepackage[normalem]{ulem}
\usepackage{caption}
\usepackage{tabularx}
\usepackage{enumitem}
\usepackage{verbatim}
\usepackage{bbding}

\usepackage{hyperref}
\usepackage{algorithmic}
\usepackage{algorithm}
\usepackage{natbib}

\makeatletter

\newtheorem{proposition}{Proposition}

\usepackage{booktabs}
\usepackage{gnuplottex}
\usepackage{makecell}
\usepackage[T1]{fontenc}

\usepackage[utf8]{inputenc}

\usepackage{microtype}

\newcommand{\printfnsymbol}[1]{%
  \textsuperscript{\@fnsymbol{#1}}%
}
\makeatother

\title{DP-FP: Differentially Private  Forward Propagation for  Large  Models  
}

\author{Jian Du\footnotemark[1] \\
  Ant Group, Sunnyvale, CA \\
  \texttt{jd.jiandu@gmail.com} \\\And
  Haitao Mi\footnotemark[1] \\
  Ant Group, Sunnyvale, CA \\
  \texttt{haitaominlp@gmail.com} \\}

\begin{document}
\maketitle
\renewcommand{\thefootnote}{\fnsymbol{footnote}}
\footnotetext[1]{The authors contributed equally to this work. The code will be  released shortly.}

\renewcommand*{\thefootnote}{\arabic{footnote}} 

\begin{abstract}
When applied to large-scale learning problems, the conventional wisdom on privacy-preserving deep learning, known as Differential Private Stochastic Gradient Descent (DP-SGD), has met with limited success due to significant performance degradation and high memory overhead when compared to the non-privacy counterpart.
We show how to mitigate the performance drop by replacing the DP-SGD with a novel DP Forward-Propagation (DP-FP) followed by an off-the-shelf non-DP optimizer.
Our DP-FP employs novel (1) representation clipping followed by noise addition in the forward propagation stage, as well as (2) micro-batch construction via subsampling to achieve DP amplification and reduce noise power to $1/M$, where $M$ is the number of micro-batch in a step.
When training a classification model, our DP-FP with all of the privacy-preserving operations on the representation is innately free of gradient bias, total noise proportionally to model size, and memory issues in DP-SGD.
As a result, our DP-FP outperforms cutting-edge DP-SGD while retaining the same level of privacy, and it approaches non-private baselines and significantly outperforms state-of-the-art DP-SGD variants.
When applied to RoBERTa-large on four downstream tasks, for example, DP-FP achieves an average accuracy of 91.34\% with privacy budgets less than 3, representing a 3.81\% performance improvement over the state-of-the-art DP-SGD and only a 0.9\% loss compared to the non-private baseline but with a significantly lower privacy leakage risk.

\end{abstract}

\section{Introduction}
\label{sec:Introduction}


Large pretrained language models, such as BERT~\citep{devlin2019bert,roberta}, have gained a lot of traction in a variety of down streaming tasks due to their powerful representation learning from massive data.
Simultaneously, machine learning models are notorious for leaking information about their (potentially sensitive) training data.
For example, it is known that adversaries can use membership inference attacks~\citep{shokri2017membership} to predict whether or not a particular data record was in the training data for certain models.
\newcite{carlini2021extracting,shokri2017membership, carlini2019secret} all demonstrate how specific and sensitive training records can be extracted from pretrained models. As a result, Differential Privacy (DP)~\citep{dp_dwork,dp_dwork2}, a gold-standard way for privacy-preserving computation, is becoming increasingly important and must be guaranteed for large pre-trained language models in order to protect privacy.

Unfortunately, when applied to large models, DP has typically struggled to produce useful models, resulting in models with either weak privacy guarantees~\citep{dupuy2021efficient, qu2021privacy} or performance far below non-private baselines.
\citet{hoory2021learning, yu2021large, li2021large} have recently demonstrated separately that performance drop when using DP can be improved by fine-tuning large pretrained models with Differentially Private Stochastic Gradient Descent (DP-SGD).
During backward propagation, DP-SGD typically applies clipping operations and random noise to the gradient computed from each data record.
However, it is widely acknowledged that DP-SGD invariably degrades model performance, raising the privacy-utility trade-off issue, and that the degradation is especially severe for large-scale models like BERT.
Aside from lowering model accuracy, per-example gradient clipping significantly increases memory costs, making DP-SGD unsuitable for large model training. \citet{li2021large} propose ghost clipping method to help alleviate memory cost.
Nonetheless, gradient clipping in DP-SGD continues to be a significant source of model accuracy degradation because it introduces gradient bias into the learning process, lowering both the convergence rate and model accuracy. Furthermore, adding noise to the clipped gradient, which scales with the huge model dimension~\cite{yu2021large}, reduces DP-SGD convergence even further. As a result, there is a near 8\% performance drop when compared to the non-DP model tested on the SST-2 dataset with moderate privacy guarantees~\citep{li2021large, yu2021large}.

In order to alleviate the issues shown above, we leverage the  post-processing property~\cite[Proposition~2.1]{dwork2014algorithmic} of DP mechanism to 
take one step back from DP in backward propagation stage,
and we pay more attention to  DP in forward computation by sacrificing non-protected labels. 
We believe this because most classification labels, such as "True" and "False," are trivial, and only the data sample without labels requires protection as long as it cannot be linked to the training data, which is true if the training data is protected under DP.
We leave DP for generation tasks as our future work.
In contrast to DP-SGD for large-scale model~\citep{li2021large, yu2021large}, we propose Differentially Private Forward Computation (DP-FC) followed by an off-the-shelf  optimizer such as SGD, Adam, etc, which yields classification models with much higher accuracy while imposing no memory or computation burden on the  training. In this paper, we make the following main contributions:
\begin{enumerate}
    \item
  Unlike DP-SGD, which has a high memory cost penalty due to clipping per-example gradient, DP-FC only clips the latent representation for noise calibration during the forward propagation stage. Because an off-the-shelf optimizer can be used, DP-FC is naturally free of gradient bias and has no additional memory cost over its non-DP counterpart.
    
\item The amount of noise required no longer scales with model size and is only related to the representation dimension.
More importantly, we propose a novel micro-batch structure that is compatible with the DP-FP process and further reduces noise power to $\frac{1}{M}$, where $M$ is the number of micro-batches in each step. The maximum amount of noise power reduction occurs when the value of $M$ is set to the size of the mini-batch.
\item We show that the DP-FP  significantly improves the model performance with strong privacy guarantee. For example, we compare DP-FP to the cutting-edge DP-SGD method for 
different privacy budgets, i.e., $\epsilon<3$, on RoBERTa-large, and it shows that  DP-FP achieves an average accuracy of 91.34\% on four downstream tasks, which is within 0.9\% of the non-private baseline and 3.81\% better than the cutting-edge DP-SGD method.
\end{enumerate}

\section{ML Model under DP}
\label{sec:formulation}
\subsection{DP Preliminary}
\label{sec:pre}
\noindent{\textbf{DP Definition}}:
DP stabilizes the output and introduces ``noise'' or random information into the process of computing the output of a function, such as an ML model, making it nearly impossible for the adversary to determine whether or not a specific data record, such as a sentence in the training data, was used.
Individual data records face the same level of privacy risk whether or not they are included in the computation due to DP assurance.

The rigorous DP definition is as follows.
Let $X$ and $X^\prime$ be neighboring datasets, i.e., $X\sim X^\prime$, such that one can be obtained from the other by either adding or removing a data record~\citep{dwork2006calibrating}.
In an NLP task, for example, $X$ and $X^\prime$ could be two datasets with sentences, 
and $X\sim X^\prime$ indicates that they differ by only one sentence.
A randomized algorithm $\mathcal M$ is $(\epsilon, \delta)$-DP if for any subsets $Y$ of $\mathcal M$, and all neighboring datasets $X \sim X^{\prime}$, we have 
\begin{equation}
    \label{eq:dp-def}
    \Pr[{\mathcal M}(X)\in {Y}]\leq\mathrm{e}^{\epsilon }\Pr[{\mathcal M}(X^{\prime})\in {Y}]+\delta.
\end{equation}
The probability distributions of the outputs $\mathcal M(X)$ and $\mathcal M(X^\prime)$ differ only by a multiplicative factor $\mathrm{e}^{\epsilon }$ and a very small value $\delta$.
Thus, the value of $\epsilon$ determines the strength of the privacy guarantee: the lower the value of $\epsilon$, the more privacy is guaranteed, and $\delta$ can be interpreted as the likelihood of failing to achieve DP.
Note that $\mathcal M(X)$ can be any complicated random function of $X$. For instance, it could be a gradient estimate in DP-SGD as shown later in~Eq.~(\ref{eq:dpsgd}) or a token/latent representation computed by neural networks in DP-FP as shown later in~Eq.~(\ref{eqn:dp-fp}).
To quantify privacy protection, data administrators impose a privacy loss cap, also referred to as a privacy budget, i.e., $(\epsilon, \delta)$, to calibrate the amount of random noise required to design $\mathcal M(\cdot)$.

\noindent {\textbf {DP Mechanism}}:
To achieve the $(\epsilon,\delta)$-DP protection defined in Eq.~(\ref{eq:dp-def}), the random function $\mathcal M(\cdot)$ must be specified.
A simple but effective method for achieving  DP protection of $X$ given $f(X)$ is obtained by first constraining the range of $f(X)$ by a clipping operator, i.e., $\mathrm{Clip}(f(X); C)$, and then randomizing the result with calibrated Gaussian noise~\citep{abadi2016deep,li2021large}. 
We introduce details of $\mathrm{Clip}(\cdot; C)$ in Eq.~(\ref{eqn:clipping-dpsgd}) and Eq.~(\ref{eqn:clipping-dpfp}) with concrete examples.
The clipping threshold $C$ is also known as {\it{sensitivity}} that 
guarantees the greatest output variation  for any pair of  $X\sim X^{\prime}$ in terms of the $\ell_2$-norm.
Mathematically, the DP mechanism is defined by
\begin{equation}
   \mathcal M(f(X))=\mathrm{Clip}(f(X); C)+\mathcal{N}(0, \sigma^2)
\end{equation}
with $\sigma^2$ being calibrated  according to  the DP profile $(\epsilon(\delta), \delta)$ given by~\citep{balle2018improving}:
\begin{equation}
\label{eqn:compute_eps}
\delta(\epsilon ; \mu)
=\Phi\left(-\frac{\epsilon}{\mu}+\frac{\mu}{2}\right)-\mathrm{e}^{\epsilon} \Phi\left(-\frac{\epsilon}{\mu}-\frac{\mu}{2}\right),
\end{equation}
where 
\begin{equation}
\label{eqn:mu}
    \mu = C / \sigma,
\end{equation} 
and $\Phi(t)$ is the c.d.f. of Gaussian distribution. 

\noindent{\textbf{Post Processing Property}}: A key DP property is its immunity to {\it{post-processing}}~\cite{dwork2014algorithmic}, which states that a differentially private output, i.e., $\mathcal M(f(X))$, can be arbitrarily transformed using some data-independent function without compromising its privacy guarantees.

\noindent{\textbf{DP Accounting}}:
When each DP mechanism meets certain DP guarantees individually, privacy suffers as a result of the composition of these DP mechanisms. Subsampling used before a private mechanism, on the other hand, increases the guarantee of privacy. Because differentially private model training entails a series of composition of subsampling and updates, DP accounting for such a series operations is required.
There are several widely used privacy accounting methodologies to adopt, such as moments accountant~\citep{abadi2016deep}, Re\'nyi DP (RDP)~\citep{mironov2017renyi}, Gaussian DP with central limit theorem~\citep{dong2021gaussian, bu2020deep}, and numerically composing tradeoff functions~\citep{gopi2021numerical, koskela2020computing, zhu2021optimal}.
RDP, a generalization of moments accounting, provides strict upper bounds on the actual privacy cost, whereas Gaussian DP with central limit theorem (CLT), while asymptotically exact, tends to underestimate the privacy loss, and composing tradeoff functions~\citep{gopi2021numerical} provides a more precise estimate.
We use Gaussian DP with CLT for DP analysis due to its simplicity and also report the total DP cost in the experimental analysis using the various methods listed above. To make the paper self-contained, we include more information about Gaussian DP with CLT in the appendix.


\subsection{DP-SGD Challenges in Large Models}
\label{sec:dpsgd}
DP assurance has been incorporated into deep learning  by appropriately randomizing conventional model updates to limit what can be breached from the training data when revealing the model~\citep{abadi2016deep}. DP-SGD, which randomizes the SGD-based optimizers such as SGD and Adam, is commonly used to achieve such a sanitized model parameter release while concealing individual training data record.

Unlike traditional  off-the-shelf optimizers, DP-SGD clips the $\ell_2$-norm of {\it per-sample gradient} before adding the calibrated noise to protect the input data in each step.
In more concrete terms, the DP mechanism
$\mathcal M(\cdot)$ for estimating the gradient in each $B$-sized batch is as follows:
\begin{equation}
\label{eq:dpsgd}
{g}=\frac{1}{B} \sum_{i=1}^{B} \left(\mathrm{Clip}\left(\nabla \mathcal{L}_{i}, C\right)+ \mathcal{N}\left(0, \sigma^2I_{d}\right)\right)
\end{equation}
with $d$  the model dimension and  
$\mathcal{L}_{i}$ the loss function of data record $i$. The clipping operator that shrinks the gradient
of an individual example whenever it exceeds some threshold $C$
is computed by
\begin{equation}
\label{eqn:clipping-dpsgd}
\mathrm{Clip}\left(\nabla \mathcal{L}_{i}, C\right)
\triangleq 
\nabla \mathcal{L}_{i}\cdot\min\left(1, \frac{C}{\left\|{\nabla \mathcal{L}_{i}} \right\|_2}\right).
\end{equation}
The model is then updated based on  ${g}$, which is known as DP-SGD\footnote{The $\text {Optimizer}(\theta, g)$ represents the DP-enhanced version of most off-the-shelf optimizers, such as DP-Adam, which, like regular Adam, performs updates and moment accumulation with privatized gradients, i.e., $g$ in Eq.~(\ref{eq:dpsgd}). In this paper, we simply refer to the DP-version optimizers as DP-SGD.}:  
\begin{equation}
 \theta \leftarrow  \text{Optimizer}\left(\theta,  
    {g}\right).
\end{equation}
 Since ${g}$ has been DP guaranteed with an $(\epsilon, \delta)$ privacy cost in Eq.~(\ref{eq:dpsgd}), $\theta$ on the left-hand-side is also DP protected with the same DP cost as DP is immune to post-processing. 
Furthermore, for $T$-step updates, the total privacy cost can be calculated by composing $(\epsilon,\delta)$ of each steps using the methods discussed    in Section~\ref{sec:pre}'s DP Accounting paragraph. 
We denote the privacy budget by $(\epsilon_T, \delta_T)$ with $\epsilon_T>\epsilon$ and $\delta_T>\delta$.
Applying the above DP-SGD to large-scale models raises the following challenges:
\begin{itemize}
    \item DP-SGD 
    requires the computation and storage of individual gradients in each step due to the  clipping operation in Eq~(\ref{eq:dpsgd}). Because each individual gradient requires the same amount of memory as the model, storing individual gradients in DP-SGD consumes a significant amount of memory~\citep{li2021large}.
    \item The clipping operation in Eq.~(\ref{eqn:clipping-dpsgd}) inescapably introduces a significant bias~\citep{chen2020understanding} in the update direction, preventing convergence  and significantly degrading model performance in general.
    \item The total noise power scales with the model size ($d$), as shown in Eq.~(\ref{eq:dpsgd}), and degrades the model performance significantly~\citep{yu2021large}. Even with small $\sigma$, the total noise power is still huge for large models, such as BERT, where $d$ scales to 110M.
\end{itemize}

\section{DP Forward-Propagation (DP-FP)}
\label{sec:dpfp}
To address the DP-SGD issues mentioned above, we propose Differentially Private Forward Propagation, or DP-FP, which protects data privacy during forward propagation.
As a result, all off-the-shelf optimizers can now be used in the back-propagation stage without DP.
As a result, DP-FP is free of DP-SGD's flaws, such as massive memory consumption, gradient bias, and a large-dimension noise vector. 
We also propose a novel micro-batch construction method for DP amplification to further reduce the noise power in each coordinate of the added noise vector.

\subsection{Forward Propagation under DP}
Let $X$ denote the entire training dataset and $\text S(X)$ be the subsampling scheme used to build a batch for each step of model update, 
such as shuffling and sampling with/without replacement, Poisson sampling.
The latent representation is then denoted in a composition form of $h\circ \text S(X)$ 
with $h(\cdot)$ being the ML model for latent representation computation, e.g., the hidden state or pooler output of \texttt{[CLS]}\footnote{In this paper, we simply use the pooler output of \texttt{[CLS]} as our choice of $h(\cdot)$, since it is straightforward to 
integrate our DP at the outside of encoders for classification tasks, without changing any code inside encoders. Furthermore, it is much smaller in size than the transformer layers preceding the pooler.}.
Note that  subsampling schemes provide DP amplifications~\cite{wang2019subsampled,dong2021gaussian}, which reduce the amount of noise required under the same privacy budget.

To achieve the  forward propagation under DP, we first stabilize the latent representation.
Because training data is random, the output of $h\circ \text S(X)$ can vary significantly, implying that the model will vary significantly if different data records are used for training. As a result, data privacy is at risk of being compromised by a membership attack.
Thus, we constrain $h$'s output range by clipping the representation that corresponds to each data record, such as a sentence in a language model.
We clip the output of $h(\cdot)$, which shrinks the latent representation whenever its $\ell_2$ norm exceeds a certain threshold $C$, similarly to DP-SGD.
More formally, the clipped latent representation is given by  
\begin{equation}
\label{eqn:clipping-dpfp}
\mathrm{Clip}\left(h(\cdot), C\right)
\triangleq 
h\cdot\min\left(1, \frac{C}{\|h(\cdot)\|_2}\right).
\end{equation}
The clipping operation also implies that the greatest variation output for a pair of neighboring datasets in terms of the $\ell_2$-norm is given by
\begin{align*}
    C = \underset{ X \sim  X^{\prime} }{\max}||h\circ \text S(X)-h\circ\text S(X^{\prime})||_2.
\end{align*}
Following clipping, Gaussian noise is added to ensure DP, with details on noise power $\sigma^2$ calibration provided later in this subsection. Before the downstream classification layer, the latent representation is computed as follows:
\begin{equation}
\label{eqn:dp-fp}
  \mathcal M(X)\triangleq  \mathrm{Clip}\left(h\circ \text S(X), C\right) + \mathcal{N}\left(0, \sigma^2I_{k}\right).
\end{equation}
Since the input data is under DP according to Eq.~(\ref{eqn:dp-fp}), the model updated based on the result of Eq.~(\ref{eqn:dp-fp}) still follows the same DP assurance according to the post-processing property of DP~\cite[Proposition~2.1]{dwork2014algorithmic}. 
As a result, after Eq. (\ref{eqn:dp-fp}), a conventional SGD-based optimizer, such as Adam, can be used and the privacy of $X$ is still guaranteed.

Until now, we identify the advantages of DP-FP over DP-SGD in the following propositions:

\noindent  

\begin{proposition}
\label{prop:noise-dim}
\textbf{Small noise dimension in DP-FP}:
It is worth noting that $I_k$ represents a $k$-dimension identity matrix, and therefore the noise vector dimension $k$ in DP-FP is much smaller than $d$, which is the noise vector dimension in Eq.~(\ref{eq:dpsgd}) that equals to the model dimension in DP-SGD. 
Take BERT for example, $d\approx$ 110M, whereas $k$ is only 768 if using the hidden state of \texttt{[CLS]} for downstream task. Thus, DP-FP saves significant total noise power than that for DP-SGD due to a significant noise dimension reduction. 
\end{proposition}
\begin{proposition}
\label{prop:unbiased-g}
\textbf{ Unbiased gradient in DP-FP}: The results of Eq.~(\ref{eqn:dp-fp}) are then fed to the classifier, 
which predicts label distribution, computes the loss further, and then performs standard backpropagation via an off-the-shelf optimizer such as Adam for model update.
As a result, the DP-FP backpropation inherits all of the advantages of the non-DP optimizer and produces an unbiased estimate of the true gradient.
\end{proposition}

\subsection{Micro-batch for DP Amplification\footnote{The micro-batch construction in our DP-FP is used to achieve DP amplification in order to reduce calibrated noise variance without sacrificing privacy, which is distinct from the micro-batch functionality used in Tensorflow Privacy to reduce the memory cost of the DP-SGD implementation.} in DP-FP} 
Intuitively, privacy amplification by subsampling is caused by the fact that individual data record has complete privacy if it is not included in the subsample. Based on this intuition, DP-SGD benefits from batch subsampling for DP amplification \citep{li2021large, yu2021large}, which reduces the calibrated noise power for each coordinate significantly.

In contrast to DP-SGD, we show that additional DP amplification can be achieved by subsampling out $M$ micro-batches that comprise a batch, resulting in lower noise power for each coordinate of the $k$-dimension noise vector in Eq.~(\ref{eqn:dp-fp}). Because of the unique structure for DP operations in forward propagation, this DP amplification is unique to DP-FP and does not exist in DP-SGD, as discussed further below.

More concretely, an independent Bernoulli trial for all data records, i.e., sentences in a dataset, is performed to construct each micro-batch with subsampling probability $p$.
Clipping is applied to each latent representation corresponding to the input data record, i.e., clip the hidden state of \texttt{[CLS]} in the BERT model. 
In the following, we evaluate the privacy cost using the Gaussian DP (GDP) framework~\citep{dong2021gaussian}, which measures the privacy profile $(\epsilon, \delta)$ in terms of $\mu = C/\sigma$ using Eq.~(\ref{eqn:compute_eps}) and (\ref{eqn:mu}). 
To make our paper self-contained, we include a preliminary of GDP calculation in Appendix~\ref{app:GDP}, as well as a more detailed procedure for privacy accounting described below.

\begin{table*}
\centering
\scalebox{0.9}{
    \begin{tabular}{c|c c c c }
    \hline
     & DP-SGD & DP-FP & RGP & SGD    \\
    \hline
     Memory cost &  $\mathcal O(md)$   & $\mathcal O(d)$   &
      $\mathcal O(m r w)$                  &
    $\mathcal O(d)$    \\
    Computational cost & 
    $\mathcal O(md)$  & $\mathcal O(md)$   
    & $
\mathcal{O}\left(m d+K r d+K r^{2}w\right)
$  & $\mathcal O(md)$  \\
    Coordinates \# to add noise & $\mathcal O(d)$  & $\mathcal O(k)$  &  $\mathcal O(w)$ & --  \\
    Noise power at each coordinate  & $\sigma^2$ & $\frac{1}{M}\sigma^2$  &$\sigma^2$  &--  \\
    \hline
    \end{tabular}}
    \caption{For all methods, $m$ is the batch size, and $d$ is the model size. In our DP-FP, $k$ and $M$ are the latent representation dimension used for downstream tasks and micro-batch number, respectively. Note that $k\ll d$, e.g. in the experiment $k=768$ while $d=3$ Million for BERT model. Specifically for RGP in~\citet{yu2021large}, $w$ is the model width, $r$ is the reparametrization rank, and
$K$ is the number of power iterations.\protect\footnotemark{}}
    \label{table:cmp}
\end{table*}

According to the subsampling DP amplification in the Gaussian DP framework~\citep{bu2020deep},
the privacy cost corresponding to each micro-batch is given by
\begin{equation}
p\cdot  G_{\mu}+(1-p) \text { Id },
\end{equation}
where  $
G_{\mu}$ is a function of $\mathcal{N}(0,1)$ and $\mathcal{N}(\mu, 1)$  with $\mu = C/\sigma$
and  $\text { Id }(\alpha)=1-\alpha$.
The details of function $
G_{\mu}(\alpha)$ is given in the appendix. 
In each step, the training stage executes $M$ micro-batches and updates $T$ steps based on the training dataset. Even if each micro-step is DP protected with a privacy cost of $(\epsilon, \delta)$, the question is whether all $T\times M$ micro-batches are private when combined, and if so, how privacy degrades as the number of steps increases, a process known as DP composition.
We have the total privacy cost
according to the central limit theorem for $T\cdot M$ rounds  Gaussian DP composition given by 
\begin{equation}
\label{eqn:gdp-clt-dpfp}
\mu_{\text{tot}}
  = p\cdot \sqrt{T\cdot M
 \left(\mathrm{e}^{(C/\sigma)^2}-1\right)},
\end{equation}
with details provided in the appendix. It is evident that the smaller $p$ is, the smaller the total privacy cost denoted by $\mu_{\text{tot}}$.

Similarly in DP-SGD~\cite{li2021large} with subsampling probability $\widetilde p$ to construct the mini-batch, the total privacy cost is given by 
\begin{equation}
\label{eqn:gdp-clt-dpsgd}
\widetilde \mu_{\text{tot}}
  = \widetilde p\cdot \sqrt{T
 \left(\mathrm{e}^{\widetilde \mu^2}-1\right)},
\end{equation}
where $\widetilde \mu = C/\widetilde \sigma$. 
To make a fair comparison, DP-FC is set to have the same batch size in expectation as DP-SGD by setting $\widetilde  p=p \cdot M$.
In the strong DP regime, $\mu$ and $\widetilde \mu$ are very small positive values close to zero. Thus, by taking the Taylor series expansion of the exponential function in (\ref{eqn:gdp-clt-dpfp}) and (\ref{eqn:gdp-clt-dpsgd}), respectively and taking into account the fact that
$
\mu_{\text{tot}} = \widetilde \mu_{\text{tot}}
$ for the same privacy budget,
we obtain $$   \frac{\sigma^2}{\widetilde\sigma^{2}}= \frac{1}{{M}}.$$
As a result, we have the third advantage of DP-FP over DP-SGD:  

\begin{proposition}
\label{prop:noise-power}\textbf{  As with DP-SGD, DP-FP requires less than $1/M$ noise power per coordinate.}
Note that the comparison is
under the same batch-size,  privacy budget, and clipping value. However, as demonstrated in~\citet{li2021large}, larger batch sizes improve performance in their DP-SGD. As demonstrated later in the experiment, our DP-FP, on the other hand, prefers small batch sizes. Due to the fact that each batch is constructed by randomly sampling each data record, the batch size decreases as $p$ decreases. Thus, $\sigma^2$ is even smaller than $\frac{1}{M}\widetilde{\sigma}^2$ according to Eq.~(\ref{eqn:gdp-clt-dpfp}).
\end{proposition}

Finally, we summarize the DP-FP algorithm in Algorithm~\ref{alg:DP-FP}, including the micro-batch construction for DP amplification.
Since the noise power in each step is calibrated according to the
DP budget of $(\epsilon, \delta)$ and total steps $T$, $(\epsilon, \delta)$
is spent out after $T$ steps.

\begin{algorithm}[t]
\caption{DP-FP Training}
\label{alg:DP-FP}
\begin{algorithmic}[1]
\REQUIRE  DP budget $(\epsilon,\delta)$, sampling rate $p$,    clipping threshold $C$, and representation $h(\cdot)$. 
\STATE Put $(\epsilon, \delta)$ into (\ref{eqn:compute_eps}) and compute $\mu$ as $\mu_{\text{tot}}$.
\STATE Calibrate noise power $\sigma^2$ by substituting $\mu_{\text{tot}}$, $T$, $M$, and $C$ into~(\ref{eqn:gdp-clt-dpfp}).
\FOR{$t=1, \ldots, T$}
    \FOR[$\triangleright$ $M$ micro-batches in each step:]{$m=1, \ldots, M$ in parallel}
        \STATE Sample micro-batch $x_m$ with Bernoulli trail for each data record 
        \STATE  $\widetilde  x_m \gets \text{Clip}\left(h(x_m);C\right) +  \mathcal{N}(0, \sigma^2 I_k)$
        \STATE $w\gets  \text{optimizer}(\widetilde  x_m, w )$ \COMMENT{$\triangleright$ SGD-based off-the-shelf optimizer for model update}
    \ENDFOR
\ENDFOR
\end{algorithmic}
\end{algorithm}

\subsection{Comparison to existing methods}
In contrast to SGD, DP-SGD necessitates the computation and storage of individual gradients, resulting in batch- and model-size-dependent memory costs. Furthermore, the total noise power in DP-SGD scales linearly with model size. As a result, applying DP-SGD to large-scale models is difficult. 

\citet{li2021large} propose ghost clipping, a memory-saving technique that allows clipping in DP-SGD to run without instantiating per-example gradients. By extending the method in \citet{lee2020scaling}, ghost clipping consumes nearly the same memory as non-private SGD training while doubling the computation throughput via numerical validation. As a result, we use the memory and computational cost of SGD as a reference for ghost clipping.
We also compare the costs of the most recent memory efficient method, reparameterized gradient perturbation (RGP)~\citep{yu2021large}. While RGP is faster per update, it requires more than three times as many epochs as the ghost clipping method as pointed out by~\citet{li2021large}.
\footnotetext{We use slightly different symbol notion for complexity analysis from that in~\cite{yu2021large} without assuming  the weight matrix is square as that in ~\cite{yu2021large}.}

\section{Experiments}
\label{sec:Experiments}
\setlength\tabcolsep{3pt}
\begin{table*}
\centering
\scalebox{0.85}{
\begin{tabular}{lccccccccccc}
\toprule
\multicolumn{1}{l}{\multirow{2}{*}{Model}}
& \multicolumn{1}{l}{\multirow{2}{*}{ Method}}
& \multicolumn{5}{c}{$\epsilon = 3$ (RDP)}  & \multicolumn{5}{c}{$\epsilon = 8$ (RDP)}  \\\cline{3-7}\cline{8-12}
\multicolumn{2}{c}{}                  & MNLI-(m/mm) & QQP & QNLI & SST-2 & Avg.   &  MNLI-(m/mm) & QQP & QNLI & SST-2 & Avg. \\
\hline
\multirow{4}{*}{BERT-base} 
& Non-DP        & 83.97/84.46 & 90.99 & 90.09 & 92.55 
& 88.41
& 83.97/84.46 & 90.99 & 90.09 & 92.55 
&88.41\\ 
& DP-SGD$^*$    & --          & --    & --    & --    &-- & 54.6/53.4   & 74.5  & 63.6  & 82.3  
&65.68\\
& RGP$^*$       & --          & --    & --    & --    &-- & 79.1/78.0   & 84.8  & 86.2  & 91.5  
&83.92\\
& Ours DP-FP    & 81.93/82.11 & 89.47 & 89.25 & 90.14 
&86.58
& \textbf{82.54/82.39} & \textbf{89.68} & \textbf{88.82} & \textbf{91.51} 
&\textbf{86.99}
\\
\hline
\multirow{3}{*}{RoBERTa-base} 
& Non-DP & 87.33/87.08 & 91.07 & 91.61 & 94.27 
&90.27
& 87.33/87.08 & 91.07 & 91.61 & 94.27 & 90.27\\ 
& DP-SGD$^+$ & 82.47/82.10 & 85.41 & 84.62 & 86.12 
&84.14
& 83.30/83.13 & 86.15 & 84.81 & 85.89 
&84.66
\\
& RGP$^+$& --          & --    &.   -- & --    & --          & 80.5/79.6   & 85.5 & 87.2  & 91.6  
&84.88\\
& Ours DP-FP  & \textbf{85.84/86.18} & \textbf{89.92} & \textbf{90.35} & \textbf{93.23} 
&\textbf{89.10}
& \textbf{86.00/85.81} & \textbf{89.84} & \textbf{90.96} & \textbf{93.23} 
&\textbf{89.17}
\\
\hline
\multirow{3}{*}{RoBERTa-large} 
& Non-DP & 90.12/89.72 & 91.60 & 93.56 & 96.22 
&92.24
& 90.12/89.72 & 91.60 & 93.56 & 96.22 & 92.24\\ 
& DP-SGD$^+$ & 85.53/85.81 & 86.65 & 88.94 & 90.71 
& 87.53
& 86.28/86.54 & 87.49 & 89.42 & 90.94 
&88.13\\
& RGP$^+$&--&--&--&--&--&86.1/86.0&86.7&90.0&93.0 
&88.36\\
& Ours DP-FP & \textbf{89.27/89.08} & \textbf{90.68} & \textbf{92.59} & \textbf{95.07} 
&\textbf{91.34}
& \textbf{89.70/89.05} & \textbf{90.81} & \textbf{93.11} & \textbf{95.18} 
&\textbf{91.57}\\
\hline
\hline
\multicolumn{2}{l}{$\epsilon \approx $ (Gaussian DP + CLT)} & 2.52 & 2.52 & 2.00 & 1.73 & -- & 5.83 & 5.85 & 4.75 & 4.33 &--\\
\multicolumn{2}{l}{$\epsilon \approx $ (Compose tradeoff func.)} & 2.75 & 2.75 & 2.57 & 2.41 & --& 7.15 & 7.16 & 6.87 & 6.69 &--\\
\bottomrule
\end{tabular}
}
\abovecaptionskip=5pt
\caption{Accuracy scores of different methods on development sets. The
DP-SGD$^*$  and RGP$^*$ results are from ~\newcite{yu2021large}; and the DP-SGD$^+$ and RGP$^+$ results are the documented number in~\newcite{li2021large}.
Best accuracy scores for each privacy level are in bold. ``Avg.'' shows average scores of four tasks.\protect\footnotemark{}}
\label{table:main_results}
\end{table*}

\subsection{Data and Settings}
Following \newcite{li2021large}, we mainly focus on fine-tuning large pretrained models on classification tasks, 
including MNLI, QQP, QNLI, and SST-2 from the GLUE benchmark~\cite{wang2018glue} that each has more than 60k training instances.
We provide the data statistics for the four tasks in Appendix~\ref{app:statistics}.

Our non differential privacy (Non-DP) baselines are finetuned BERT-base, RoBERTa-base, and RoBERTa-large. 
For classification task, following common settings, we add a linear layer over the output of the pooler layer for encoder, 
the pooler layer simply takes the hidden state of \texttt{[CLS]} token as input, and applies another linear dense layer and a $tanh(\cdot)$ activation.
We train our baselines on four data sets for 5 epochs with a learning rate $2\times 10^{-5}$, a batch size $32$, and a max input length $128$.
We save and validate each model for each epoch, and report the best accuracy scores as our baseline systems.

During the fine-tuning stage, our DP-FP method adds noise and clipping operations (as in Algorithm~\ref{alg:DP-FP}) before the linear classification layer based on the following two facts. 
First this approach treats pretrained model encoder classes as black boxes and does not require any code change inside encoder classes.
Second, typically for large-pretrained models, the noise dimension, i.e., 768 for BERT-base and RoBERTa-base, and  1024 for RoBERT-large, is fixed and much smaller than the model size, i.e., 110M for BERT-base.
Then we apply standard AdamW~\cite{adamw} optimizer in the back-propagation stage.
Please note that we do not add any noise and clippings in the inference time, as we only need to protect the fine-tuning training data.
In our following experiments, we use the following hyper-parameters as our default settings for our DP-FP method: 
total fine-tuning epoch is three,  $M = 32$, $C=1.0$, learning rate is $5 \times 10^{-6}$,  max input length is $128$,  the micro-batch subsampling rate $p=\frac{B}{M\cdot D}$ and the expected batch size $B=32$.
We consider a practical scenario in which the total amount of privacy budget $(\epsilon, \delta)$ is constrained. For both DP-FP and DP-SGD, this privacy budget constraint corresponds to a constraint on the number of data samples used in the tuning process.
As a result, we report accuracy scores on development sets once the privacy budget has been depleted.

\footnotetext{The results for SGD$^*$ and RGP$^*$ in Table~\ref{table:main_results} are  from documented numbers in~\citet{yu2021large}. These results are under the DP guarantees of $\delta=10^{-5})$. These guarantees are strictly weaker than those provided by DP-SGD$^+$ and our DP-FP, which are based on $\delta = 1/2D$ (note that the smallest dataset in these tasks contains $D>60$k records).}

To ensure a fair comparison with the DP-SGD method in large-scale models reported in the literature, we set the same privacy budget for Gaussian DP with CLT as that in~\citet{li2021large} for each experiment using $\epsilon$ in the set $\{1.73, 2, 2.52,   4.33, 4.75, 5.83, 5.85\}$ and $\delta = 1/2D $ with $D$ the data record number in the training set.
We then numerically calibrate $\sigma^2$ as in Line 2 of Algorithm~\ref{alg:DP-FP}.
We also report the corresponding total privacy cost documented in~\citep{li2021large} calculated by RDP and composing tradeoff function methods. Please see Section~\ref{sec:pre} and the references therein for further information.

\subsection{Main Results}
Table~\ref{table:main_results} shows the main results on four tasks. 
The larger the pretrained models, the higher the accuracies for the non-DP baselines, ranging from BERT-base (110M parameters) to RoBERTa-large (355M parameters).

We compare full fine-tuning with reparameterized gradient perturbation
(RGP)~\citep{yu2021large}, and memory efficient
DP-SGD for large language model~\citep{li2021large}
as they are the state-of-the-art methods for DP fine-tuning on sentence classification at
the time of writing. 
The DP-SGD scores are documented in \newcite{li2021large}, and DP-SGD hurts the baseline performance by 4-6$\%$ in average for RoBERTa-base and RoBERTa-large, even with a larger $\epsilon$ at 8 (RDP).
In particular, results of RoBERTa-base models show that DP-SGD still degrades performance by up to 8$\%$ for the SST-2 data set.  
By contrast, RGP of \newcite{yu2021large} reduces the gap significantly to 1$\%$ point on SST-2 data set, but is still at least 4$\%$ 
far behind Non-DP BERT-base on other three tasks.

We run DP-FP on each level  of the privacy budget for three epochs of steps, and report the final scores only for the model reaches the privacy budget. Note that, we use the default setting and same hyper-parameters as illustrated in the previous subsection for each experiment of our DP-FP in  Table~\ref{table:main_results}.
As shown in Table~\ref{table:main_results}, DP-FP significantly improves accuracy over existing methods and closes the gap to the Non-DP counterpart. It has only a 0.58$\%$ performance drop for SST2 on RoBERTa-large with $\epsilon$ set to 1.73 under GDP with CLT in particular. We further average this DP-FP  performance drops  across datasets  for each of the models. When $\epsilon=3$ (RDP), the average performance drops to Non-DP model are within 0.83-1.78$\%$; and when $\epsilon=8$, they are within 0.68-1.43$\%$. 
Moreover, DP-FP is clearly better than DP-SGD and RGP approaches.
This significant performance advantage stems in part from the fact that DP-FP overcomes the biased gradient problem in DP-SGD 
and also has a lower noise power due to micro-batch number as well as a lower  noise dimension.

More interestingly,  the larger the pretrained model, the smaller the gap to the Non-DP model for DP-FP. 
For example, on the SST-2 task for $\epsilon=3$ (RDP), the gap between  DP-FP and Non-DP is reduced from 2.41$\%$ (BERT-base)  to 0.58$\%$ (RoBERTa-large).
The main reason for this, we believe, is that because our DP-FP adds noise to the latent representation before the linear classification layer, the total noise power does not scale with model size. 

\begin{figure}
    \centering
    \includegraphics[width=.45\textwidth]{"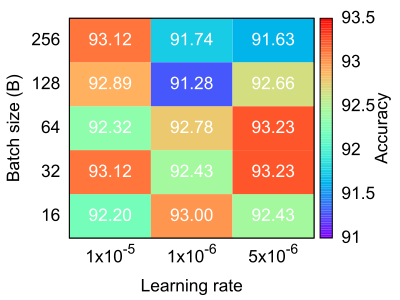"}
\abovecaptionskip=2pt
\caption{Accuracy scores for different batch sizes ($B$) on SST-2 development set.}
\label{fig:batch_heatmap}
\end{figure}

\subsection{Hyperparameter Tuning}

As suggested in~\newcite{li2021large}, DP optimization is sensitive to hyper-parameter selection.
We therfore examine the performance of DP-FP for various hyper-parameters, as our DP-FP differs significantly from DP-SGD in terms of architecture.
In this section, we report all results of RoBERTa-base with DP-FP on SST-2 data set, and we fix the total training epochs to be 3 to deplete the privacy budget, i.e., $(\epsilon,\delta)=(3,1/2D)$.

Figure~\ref{fig:batch_heatmap} shows the heat map of accuracy scores for different batch sizes with 
$M = 1$ and $C = 1.0$. 
Those results suggest that 1) large batch sizes, like 128 and 256, hurt performance; 
2) DP-FP requires a small learning rate in order to achieve better performance. 
Those two interesting observations are opposite of the findings of DP-SGD in~\newcite{li2021large}.
The intuitive explanation is given below. 
By substituting $p=B/(M\cdot D)$ into Eq.~(\ref{eqn:gdp-clt-dpfp}), we have the privacy parameter
\begin{equation}
    \label{eqn:exp}
\mu_{\text{tot}}
  = \frac{B}{M\cdot D} \sqrt{T\cdot M
 \left(\mathrm{e}^{(C/\sigma)^2}-1\right)}.
 \end{equation}
 Intuitively, this demonstrates that, given a fixed privacy budget, which implies a fixed $\mu_{\text{tot}}$, noise power $\sigma^2$ decreases as batch-size $B$ decreases. However, we cannot reduce $B$ too much because the model will not converge.

\begin{figure}
    \centering
    \includegraphics[width=.45\textwidth]{"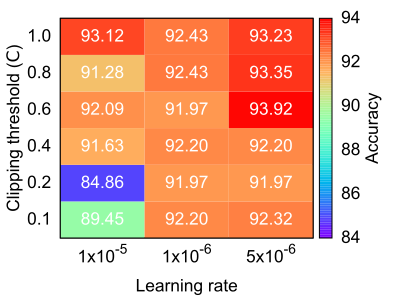"}
\abovecaptionskip=2pt
\caption{Accuracy scores for different clipping threshold ($C$) on SST-2 development set.}
\label{fig:clip_heatmap}
\end{figure}


\begin{figure}
    \centering
    \includegraphics[width=.45\textwidth]{"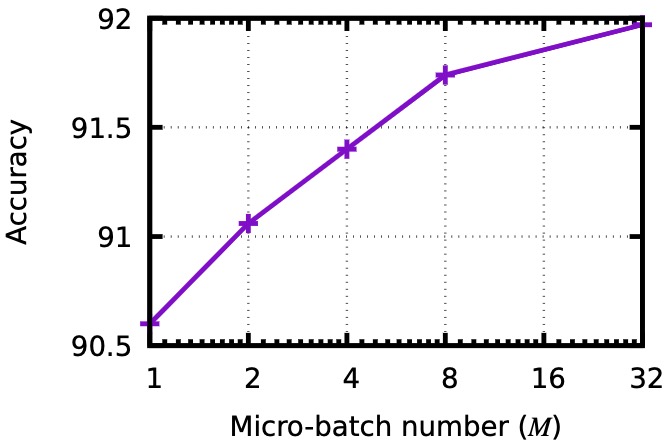"}
\abovecaptionskip=2pt
\caption{Accuracy curve for different micro-batch numbers ($M$) on SST-2 development set.}
\label{fig:micro_batch_curve}
\end{figure}

Figure~\ref{fig:clip_heatmap} shows the heat map of accuracy scores for different clipping threshold ($C$) with $\epsilon = 3$, 
$M = 1$, and $B = 32$.
\newcite{li2021large} show that small clipping thresholds lead to the best performance by setting $C=0.1$.
In contrast, we find different trends from that in~\newcite{li2021large} such that smaller clipping thresholds lower than 0.4 actually hurt
performance, and there is no significant changes in larger clipping thresholds (from 0.6 to 1.0). 
Removing too much of the latent representation with a small $C$ in DP-FC results in too much loss of semantic information and leads to significant performance drop. 

Finally, Figure~\ref{fig:micro_batch_curve} shows the curve of DP-FP with different micro-batch numbers on SST-2 development set.
In this experiment, we set the privacy budget $\epsilon$ to be 0.02 under Gaussian DP + CLT, a very strong privacy level, the batch size to be 32, and the total number of epoch to be 3.
Larger micro-batch numbers clearly lead to better performance, as the larger the $M$, the more noise power is reduced (by $\frac{1}{M}$), as shown in Proposition~\ref{prop:noise-power}.

\section{Related Work}
\label{sec:RelatedWork}
In the research line of sharing models while protecting the corresponding training data privacy, 
previous work mainly use DP-SGD to train privacy-preserving models~\citep{shokri2015privacy, yu2019differentially}.
\citet{abadi2016deep} propose a tight privacy accounting leading to a reasonable level of privacy cost for DP-SGD.
Following that, variants of DP-SGD have been proposed to improve the model accuracy such as clipping based on quantiles~\cite{andrew2019differentially}, dynamic noise power and clipping value~\cite{du2021dynamic}, 
and dynamic learning rate~\cite{wu2021adaptive}, etc.
DP-SGD and its variants, on the other hand, have had limited success in large deep learning models due to their high computation cost, huge memory overhead, and significant performance drops.

In order to reduce the huge memory cost of each DP-SGD step, 
\citet{subramani2020enabling, anil2021large} study an effective use of JAX primitives in
conjunction with the XLA compiler. 
Later \citet{li2021large} mitigate the memory cost by a ghost clipping method, but still has 2 times computation throughput cost due to per-example grade clipping.
\citet{yu2021large} reduce the memory cost via rank decomposition.

To improve the model accuracy, 
\newcite{anil2021large} successfully implement mega batch size, e.g., a batch-size of 2M, for DP-BERT for model accuracy improvement, but still has a 10\% accuracy drop compared to the non-private BERT model.
\citet{dupuy2021efficient} investigate private BERT fine-tuning, but reported results with $\epsilon$ at least 100.
Very recently, studies in \citet{hoory2021learning, yu2021large, li2021large} show that the DP-SGD's model performance drop can be mitigated by the use of large pretrained models. Event though, there is still an obvious performance gap to the non-DP counterpart due to gradient bias and additive noise.

\section{Discussion}
It is instructive to compare the types of information that DP-SGD and DP-FP protect under DP.
As illustrated in Eq.~(\ref{eqn:clipping-dpsgd}), in the DP-SGD, the label information is embedded in the gradient, 
thus the data record under DP protection is each pair of training data samples and their labels.
While in the DP-FP, as illustrated in Eq.~(\ref{eqn:clipping-dpfp}), because the DP operation is performed in the forward stage, the data record under DP protection is simply the training data sample without labels.
But, it is interesting to note that for the majority of classification tasks, one only needs to protect the privacy of the data sample, as the labels themselves are finite and not constitute privacy information as long as they cannot be connected to the training data sample.
Consider the SST-2 sentence classification task as an illustration, which contains 67,500 sentences that require protection under DP. 
Additionally, each sentence is labeled with ``positive'' or ``negative''. 
Because DP-FP ensures that the adversary almost never recovers any sentence using the fine-tuned model, labels cannot be associated with the training sentences.
However, in DP-FP for the generation task, protecting the label information is required and difficult, and further model architecture design is required. We plan to investigate it as part of our future work.

\section{Conclusion}
In this paper, we have introduced the differentially private forward propagation (DP-FP) method for applying differential privacy to large pretrained models on classification tasks. 
The key design of DP-FP exploits differential privacy's post-processing property, ensuring privacy by protecting the latent representation in the forward stage rather than the conventional wisdom of DP Stochastic Gradient Descent (DP-SGD), which protects the gradient in the backward stage.
DP-FP has the same memory cost as an off-the-shelf SGD-based optimizer, an unbiased gradient, and significantly lower noise power that scales only with the latent representation dimension, as opposed to DP-SGD, which has a large memory cost, a biased gradient, and total noise power that scales with the huge model size.
We have also created micro-batches that are unique to DP-FP in order to reduce the noise power for each coordinate.
As a result, on a large model like RoBERTa-large, DP-FP achieves an average accuracy of 91.34\% on four downstream tasks with $\epsilon$ less than 3, which is only within 0.9\% lower than  the non-private baseline and 3.81\% better than the state-of-the-art DP-SGD method.

\bibliography{anthology,custom}
\bibliographystyle{acl_natbib}

\clearpage
\appendix

\section{Supplementary Formalism Details}
\label{sec:appendix}
\begin{table*}[b]
    \centering
    \begin{tabular}{c|c|ccc}
    \hline
    Dataset & Task Type & Classes & Average Length & Training Sample Size (D) \\
    \hline
    MNLI & Natural Language Inference & 3 & 32.96 & 240,942\\
    QQP & Semantic Matching & 2 & 24.77 & 384,348\\
    QNLI & Question Answering & 2 & 39.74 & 104,374 \\
    SST-2 & Sentiment Analysis & 2 & 9.94 & 67,349 \\
    \hline
    \end{tabular}
    \caption{Statistics of four GLUE data sets.}
    \label{table:datasets}
\end{table*}

\subsection{Privacy Accounting for DP-FP }
\label{app:GDP}
Based on hypothesis testing of two Gaussian distributions, Gaussian DP define a canonical single-parameter family of privacy notions.
It gives  a computationally efficient tool for analyzing the exact composition of private algorithms in a tractable way. We present some key results in the following section, and for more information, please see~\citet{dong2021gaussian}.

\noindent{The hypothesis testing view of privacy notion}: 
Let $P$ and $Q$ denote the distributions of random function output $\mathcal M(X)$ and $M\left(X^{\prime}\right)$ with $X\sim X^{\prime}$, and let $\phi$ be any  rejection rule for testing the hypothesis $H_{0}: P$ against $H_{1}: Q$. With these in place, we can define the trade-off function between $P$ and $Q$ as follows:
\begin{equation}
\begin{aligned}
\text{T}(P, Q):[0,1] & \mapsto[0,1] \\
\alpha & \mapsto \inf _{\phi}\left\{1-\mathbb{E}_{Q}[\phi]: \mathbb{E}_{P}[\phi] \leqslant \alpha\right\},
\end{aligned}
\end{equation}
where,  $\mathbb{E}_{P}[\phi]$ and  $1-\mathbb{E}_{Q}[\phi]$  are the type I and type II errors of the rejection rule of $\phi$, respectively.
It is shown that $\text{T}(P,Q)\geq \text{T}(\mathcal N(0,1),\mathcal N(\mu,1))$. The function $\text{T}(\mathcal N(0,1),\mathcal N(\mu,1))$ a single-parameter funcation denoted by $G_{\mu}$, which is referred to as  $\mu$-GDP.

In each micro-batch of the  DP-FP in (\ref{eqn:dp-fp}) with the Gaussian mechanism, it achieves $\mu$-GDP with $\mu = \frac{C}{\sigma}$.
Consider the  sampling scheme $\operatorname{S}(X)$ that each individual data record is subsampled independently with probability $p$ from the training set to construct the micro-batch.
\citep{bu2020deep} demonstrates that given two neighboring datasets $X$ and $X^\prime$, if a random mechanism $\mathcal M$ is $G_{\mu}$-DP, then after the subsampling with $p$ denoted by $\operatorname{S}$, it holds that
\begin{equation}
\label{GDP-sampleing}
\text{T}\left(\mathcal M \circ \operatorname{S} (X), \mathcal M \circ \operatorname{S}\left(X^{\prime}\right)\right) \geqslant p G_{\mu}+(1-p) \mathrm{Id},
\end{equation}
where $\operatorname{Id}(x) = 1 - x$.
Then, after $M$ micro-batches in each step and a total of  $T$ steps, a Berry-Esseen style CLT result is indicated by \citet{bu2020deep} that as $M\cdot T\to +\infty$ and $p\sqrt{M\cdot T}\to$ a constant,  the composition of the r.h.s. of (\ref{GDP-sampleing}) converges to a $G_{\mu_{\text{tot}}}$-DP with
\begin{equation}
\label{CLT1}
        \mu_{\text{tot}} = p\sqrt{TM(e^{\mu^2}-1)}.
\end{equation}
Then the total privacy cost in terms of ($\epsilon,\delta$) after $T$ steps can be computed according to~(\ref{eqn:compute_eps}), which is repeated below:
\begin{equation}
\label{eqn:compute_eps_t}
\delta(\epsilon ; \mu)
=\Phi\left(-\frac{\epsilon}{\mu}+\frac{\mu}{2}\right)-\mathrm{e}^{\epsilon} \Phi\left(-\frac{\epsilon}{\mu}-\frac{\mu}{2}\right),
\end{equation}
where 
$
    \mu = C / \sigma.
$

\subsection{Datasets Statistics}
Please refer Table~\ref{table:datasets}.
\label{app:statistics}

\end{document}